\documentclass[runningheads]{llncs}

\usepackage{graphicx}
\usepackage{hyperref}
\usepackage{booktabs}
\usepackage{orcidlink}
\usepackage{svg}
\usepackage[figuresleft]{rotating}
\usepackage{pdflscape}
\usepackage{xurl}
\usepackage{cite}

\usepackage[flushleft]{threeparttable}
\usepackage{multirow}
\usepackage[T1]{fontenc}
\usepackage{graphicx}
\begin{document}
\title{Exploring the Effect of Dataset Diversity in Self-Supervised Learning for Surgical Computer Vision}
\titlerunning{Dataset Diversity in Self-Supervised Learning for Surgical Computer Vision}
%
\author{Tim J.M. Jaspers\inst{1}\thanks{Corresponding Author}\orcidlink{0009-0001-8306-5058}\and Ronald L.P.D. de Jong\orcidlink{0009-0005-7806-4340} \inst{2} \and Yasmina Al Khalil \inst{2}  \and Tijn Zeelenberg\inst{1}  \and Carolus H.J. Kusters\inst{1}\orcidlink{0009-0004-3114-3888} \and Yiping Li \inst{2} \and Romy C. van Jaarsveld \inst{3} \and Franciscus H.A. Bakker \inst{4,5}  \and Jelle P. Ruurda \and\inst{3} Willem M. Brinkman\inst{4} \and Peter H. N. De~With\inst{1} \and Fons van der Sommen\inst{1}\orcidlink{0000-0002-3593-2356}}


%
\authorrunning{T.J.M. Jaspers et al.}
%

\institute{Department of Electrical Engineering, Video Coding \& Architectures, Eindhoven University of Technology, Eindhoven, The Netherlands\newline \url{{t.j.m.jaspers, c.h.j.kusters, p.h.n.de.with, fvdsommen}@tue.nl} \and Department of Biomedical Engineering, Medical Image Analysis, Eindhoven University of Technology, Eindhoven, The Netherlands \and Department of Surgery, University Medical Center Utrecht, Utrecht, The Netherlands \and Department of Oncological Urology, University Medical Center Utrecht, Utrecht, The Netherlands \and Department of Urology, Catharina Hospital, Eindhoven, The Netherlands}

\maketitle              
\begin{abstract}
Over the past decade, computer vision applications in minimally invasive surgery have rapidly increased. Despite this growth, the impact of surgical computer vision remains limited compared to other medical fields like pathology and radiology, primarily due to the scarcity of representative annotated data. Whereas transfer learning from large annotated datasets such as ImageNet has been conventionally the norm to achieve high-performing models, recent advancements in self-supervised learning~(SSL) have demonstrated superior performance. In medical image analysis, in-domain SSL pretraining has already been shown to outperform ImageNet-based initialization. Although unlabeled data in the field of surgical computer vision is abundant, the diversity within this data is limited. This study investigates the role of dataset diversity in SSL for surgical computer vision, comparing procedure-specific datasets against a more heterogeneous general surgical dataset across three different downstream surgical applications. The obtained results show that using solely procedure-specific data can lead to substantial improvements of 13.8\%, 9.5\%, and 36.8\% compared to ImageNet pretraining. However, extending this data with more heterogeneous surgical data further increases performance by an additional 5.0\%, 5.2\%, and 2.5\%, suggesting that increasing diversity within SSL data is beneficial for model performance. The code and pretrained model weights are made publicly available at \url{https://github.com/TimJaspers0801/SurgeNet}.
\end{abstract}
\section{Introduction}
Over the last decade, computer vision applications in minimally invasive surgery have significantly expanded, encompassing tasks such as anatomy recognition~\cite{Mascagni2022, denBoer2023, Bakker2024}, surgical phase recognition~\cite{PADOY2012, Yitong2022}, and surgical training~\cite{HASHIMOTO2017}. Despite this growth, computer vision has not yet achieved an impact in surgery, as compared to other fields like pathology or radiology, where advanced applications are already approaching market introduction~\cite{MAIERHEIN2022}. It has been concluded that the surgical field is still in its infancy~\cite{denBoer2022, MAIERHEIN2022}, primarily due to the lack of representative annotated data~\cite{MAIERHEIN2022}.

In the broader context of computer vision, transfer learning has emerged as a powerful strategy for leveraging related datasets. Transferring pretrained weights can significantly reduce the amount of labeled data required to achieve high-performing models. Conventionally, large annotated datasets such as ImageNet~\cite{ImageNet} have been the cornerstone of transfer learning, providing state-of-the-art results for many years. However, recent advancements have demonstrated that self-supervised learning~(SSL) methods, which learn generic representations from unannotated data, can outperform supervised pretraining by providing better model initialization~\cite{caron2021}. This stems from SSL’s ability to capture a wider variety of features from diverse data, reducing the risk of overfitting to specific labeled datasets and enhancing the model's generalization capabilities.

In medical image analysis, in-domain self-supervised pretraining has offered superior performance compared to ImageNet-based initialization \cite{wang2023foundation, Hirsch2023}. A large-scale benchmark study has explored the feasibility of SSL methods within the surgical domain, offering valuable insights into methods, training settings, and frame sampling rates~\cite{RAMESH2023}. Other recent research has focused on scaling up SSL pretraining for surgical computer vision to achieve state-of-the-art performance~\cite{wang2023foundation, Hirsch2023}. The study by Alapatt~\textit{et al}.~\cite{alapatt2023} is particularly relevant to our work, since it highlights critical considerations regarding the composition of pretraining data.

The proposed research aims to delve deeper into the effects of dataset diversity in SSL for surgical computer vision. Although unlabeled data is abundantly available within this field, as most videos are recorded at over 25~frames per second~(fps) and often last several hours, the diversity within a single video is limited. Large-scale pretraining requires substantial computational resources and is therefore often expensive. This study seeks to improve understanding of the optimal dataset composition for pretraining, aiming to produce more effective and generic model weights for surgical computer vision applications. 

\section{Methodology}
\subsection{Self-Supervised learning datasets}
To analyze the effects of SSL for surgical computer vision applications, we construct SurgeNet, a comprehensive dataset comprising over 2,636,790~frames from more than seven different surgical procedures. SurgeNet includes both public and private datasets, offering a diverse collection of surgical video data. An overview of SurgeNet summarizing its contents can be found in Table~\ref{tab:surgenet}. Most datasets are publicly accessible, however, to enhance diversity/size and to create meaningful procedure-specific datasets SurgeNet is further enlarged in specific ways. We extend the public datasets with two private datasets: one from the Antoni van Leeuwenhoek Hospital~(AvL) Amsterdam, while the second dataset originates from the University Medical Center Utrecht~(UMCU), the Netherlands. All frames are extracted at a rate of 1~fps from the videos, with black edges removed.

From SurgeNet, we derive three procedure-specific subsets. The SurgeNet-CHOLEC dataset contains all frames extracted from laparoscopic cholecystectomy procedures, resulting in a total of 250,655~unique frames. The robot-assisted minimally invasive esophagectomy procedure-specific (SurgeNet-RAMIE) dataset comprises 377,287~frames. Similarly, the SurgeNet-RARP dataset includes 382,416~frames from robot-assisted prostatectomy procedures. Each of these subsets is used to analyze the impact of procedure-specific data on SSL~pretraining. Lastly, we also randomly create SurgeNetSmall including 10\% of the SurgeNet dataset, making it comparable in terms of size to the procedure-specific datasets.

\begin{table}[]
    \scriptsize
    \setlength{\tabcolsep}{4pt}
    \centering
    \begin{tabular}{l | l | l | c | c}
        \hline
        \toprule
        Procedure-           & \multirow{2}{*}{Dataset}                   & \multirow{2}{*}{Procedure}                         & \multirow{2}{*}{Incl. frames}  & \multirow{2}{*}{Public}    \\
        specific subset & & & & \\
        \midrule
        \multirow{3}{*}{CHOLEC}             & Cholec80 \cite{Cholec80}  & Laparoscopic Cholecystectomy      & 179,164       & Yes       \\
                                            & HeiChole \cite{heichole}  & Laparoscopic Cholecystectomy      & 53,427        & Yes       \\
                                            & hSDB-Chole \cite{hsdb}    & Laparoscopic Cholecystectomy      & 18,064        & Yes       \\
        \midrule
        \multirow{1}{*}{RAMIE}              & RAMIE-UMCU                & RA Esophagectomy                  & 377,287        & No        \\
        \midrule
        \multirow{3}{*}{RARP}               & ESAD  \cite{esad}         & RA Prostatectomy                  & 47,282        & Yes       \\
                                            & PSI-AVA \cite{psi-ava}    & RA Prostatectomy                  & 73,618        & Yes       \\
                                            & RARP-AvL                  & RA Prostatectomy                  & 261,516       & No        \\

        \midrule
        \multirow{6}{*}{Others}             & DSAD \cite{dsad}          & RA Rectal Resection/Extirpation  & 14,623       & Yes       \\
                                            & GLENDA \cite{glenda}      & Gynecologic Laparoscopy           & 25,682        & Yes       \\
                                            & LapGyn4 (v1.2) \cite{lapgyn}  & Gynecologic Laparoscopy           & 59,616        & Yes       \\
                                            & MultiBypass140 \cite{multibypass}     & Laparoscopic Gastric bypass surgery & 749,419        & Yes       \\
                                            & hSDB-Gastric \cite{hsdb}              & RA Gastrectomy                      & 35,576        & Yes       \\
                                            & SurgLoc2022 \cite{surgloc}            & 11 different RA porcine procedures  & 741,516        & Yes       \\
        \midrule
        \midrule
        SurgeNet                            & All of the above                      & All of the above                       &  2,636,790 & Part   \\
        \bottomrule
    \end{tabular}
    \vspace*{0.1cm}
    \caption{Composition of SurgeNet. SurgeNet comprises over 2.6 million frames derived from more than seven distinct surgical procedures. A part of SurgeNet is publicly available (over 1.9M frames).}
    \label{tab:surgenet}
\end{table}

\subsection{Downstream datasets}
The experiments in this study are conducted on three different downstream datasets, each corresponding to a specific surgical procedure: laparoscopic cholecystectomy~(LC), robot-assisted radical prostatectomy~(RARP), and robot-assisted minimally invasive esophagectomy~(RAMIE). All three datasets focus on anatomy segmentation, of which the specifications are summarized in Table~\ref{tab:downstream}. It should be noted that all videos of patients included in the test set are excluded from SurgeNet. For the CholecSeg8k dataset, all patients present in the downstream training set are also included in the SurgeNet-pertaining dataset. In the case of the RAMIE downstream dataset, 22 of the 50 patient videos are also included in the SurgeNet-pretraining dataset. There is no overlap between the patients in the RARP downstream dataset and RAPR videos in the the SurgeNet dataset.

The first dataset is the public CholecSeg8k dataset~\cite{Cholecseg8k}, derived from the Cholec80 dataset~\cite{Cholec80}, which consists of 80~videos of LC~surgeries performed by 13~surgeons. The CholecSeg8k dataset includes a selected subset of these videos, annotated with semantic segmentation masks. Specifically, 80~frames from 101~video fragments were annotated, totaling 8,080~frames. Among these, 6,800~(84\%) were used for training and 1,280~(16\%) were selected for testing. Initially, the dataset contained 13~classes: background, abdominal wall, blood, connective tissue, cystic duct, fat, gallbladder, gastrointestinal tract, grasper, hepatic vein, L-hook electrocautery, liver, and liver ligament. For this research, classes with low prevalence were excluded to ensure a more robust and reliable analysis. The excluded classes are blood, cystic duct, hepatic vein, and liver ligament.

The second dataset consists of 869~frames from 31~distinct patients undergoing RAMIE, with 749~frames~(86\%) allocated for training and 120~frames~(14\%) reserved for testing. These frames were labeled by two students with a medical background, and two research fellows in surgery or medical imaging under the supervision of an expert surgeon. In total, 12~distinct classes were annotated, including four classes for surgical tools: forceps, hook, suction \& irrigation, and vessel sealer. The other eight classes represent vital anatomical structures during RAMIE: airways (including the trachea, left main bronchus, and right main bronchus), aorta, azygos vein \& vena cava, esophagus, nerves, pericardium, right lung, and thoracic duct. The airways were grouped into a single class, due to their similar appearance and the difficulty in defining exact boundaries between the trachea and bronchi. Similarly, the azygos vein and vena cava were treated as one class for the same reason.

The third dataset focuses on anatomy recognition during RARP procedures. The total dataset consists of 282~frames from 114~patients. The training set comprised 84~procedures from distinct patients, totaling 252~(89\%) annotated frames for model training. All videos were sourced from a single-center cohort at the AvL in Amsterdam. Annotation covered four structures: the prostate, urethra, catheter, and ligated dorsal venous plexus. For the test set, one frame from 30 different patient videos was annotated. Training annotations were performed by a research fellow in the field of surgery under the supervision of an expert surgeon, while the expert surgeon solely completed test annotations. Further details on this dataset can be found in a previous publication~\cite{Bakker2024}.
\begin{table}[]
    \scriptsize
    \setlength{\tabcolsep}{4pt}
    \centering
    \begin{tabular}{l | c  c | c c | c}
        \hline
        \toprule
        \multirow{2}{*}{Downstream Dataset}           & Training  & Training  & Test     & Test  & \multirow{2}{*}{Structures included} \\
                                                      & patients  & frames    & patients & frames &   \\ 
        \midrule
        CholecSeg8k~\cite{Cholecseg8k} & 13  &    6,800     &  4   &  1,280 & 8 \\
        RAMIE                          & 27  &    749       &  4  & 120 & 12 \\
        RARP                           & 84  &    252       &  30  &  30    & 4 \\
        \bottomrule

    \end{tabular}
    \vspace*{0.1cm}
    \caption{Descriptions of the downstream datasets. The datasets exhibit diversity in terms of patient inclusion, annotated frames, and the number of structures, offering a representative sample of surgical downstream datasets that could benefit from self-supervised learning on SurgeNet.}
    \label{tab:downstream}
\end{table}

\subsection{Model architecture and Training configurations}
\textbf{Architecture:} This study emphasizes the impact of data on SSL, therefore we select a state-of-the-art architecture in the natural computer vision domain. Conventional architectures like ResNet~\cite{ResNet} and EfficientNet~\cite{EfficientNet} are CNN-based, but recent advancements have favored the Vision Transformer~\cite{VisionTransformer}. However, due to the low-resolution outputs, the vanilla Vision Transformer has difficulties regarding various dense prediction tasks~\cite{wang2021pyramid}. Therefore, we adopt the CAFormer-S18~\cite{MetaFormer} as the encoder for this study. The CAFormer-S18 builds upon the abstract MetaFormer approach, where the token mixer is not predefined, while other components remain consistent with Transformers. CAFormer-S18 incorporates convolution blocks in the first two stages and attention blocks in the last two stages, achieving state-of-the-art performance on natural image datasets like ImageNet, with 26M parameters. Concerning the downstream application, the CAFormer-S18 encoder architecture is integrated into a feature pyramid network~(FPN)\cite{FPN}. 

\noindent\textbf{Pretraining configurations: }As SSL objective, we use the well-known framework proposed by Caron et al.~\cite{caron2021}, called “Self-Distillation with NO Labels” (DINO). This method employs distillation-based techniques to enable efficient learning with smaller batch sizes, thereby reducing the demand for extensive computational resources. We closely follow the original implementation. Notably, we initiate pretraining from ImageNet-initialized weights as recommended by Ramesh et al.~\cite{RAMESH2023}, and then align with the concept of fine-tuning natural computer vision foundation models for medical applications. The pretraining of the CaFormer-S18 architecture is performed on four A100 GPUs~(NVIDIA Corp., CA, USA), each with 40GB of VRAM, using a maximum feasible batch size of~544 for 50~epochs. Note that due to computational constraints, the CAFormer-S18 network on the complete SurgeNet is only trained for 25~epochs.

\noindent\textbf{Downstream training settings: }
The downstream training process involves a five-fold cross-validation at the patient level using the training data. All training parameters are kept constant for each experiment. Frames are resized to 256$\times$256~pixels using bicubic interpolation. The cross-entropy loss is used, while the Adam optimizer is employed with a learning rate of 1$\times$10$^{-5}$. The learning rate is halved after 10~epochs without a decrease in the validation loss. All models are trained on a GeForce RTX 2080 Ti GPU~(NVIDIA Corp., CA, USA) with a batch size of 16 and an early stopping criterion of 15~epochs. Limited data augmentation techniques are used, including horizontal and vertical flipping, and rotation, each with a probability of 50\%. 

\begin{figure}[h!]
    \centering
    \includegraphics[width=\textwidth]{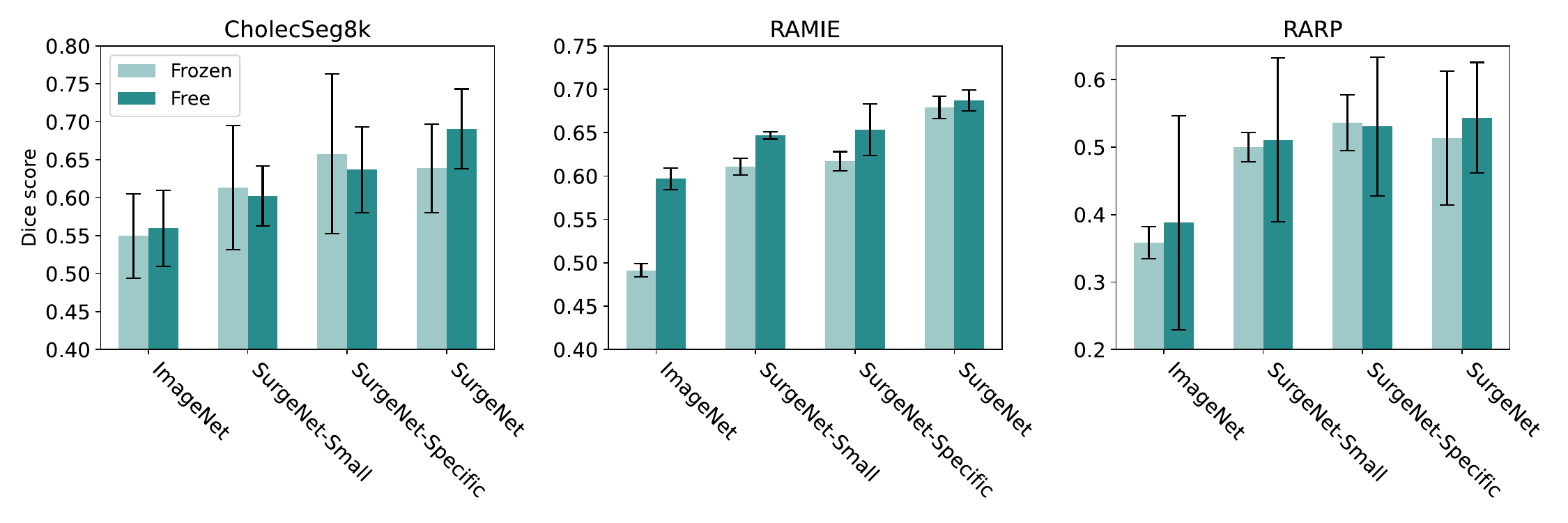}
    \caption{Segmentation results (Dice scores) on the three downstream datasets from the fivefold cross-validation. The performance is given for both fine-tuning the entire segmentation model and the model with a frozen encoder.}
    \label{fig:resultsSurgNet}
    \vspace*{-0.1cm}
\end{figure}
\section{Results and Discussion}
Figure~\ref{fig:resultsSurgNet} presents the results of SSL pretraining on the SurgeNet and its procedure-specific datasets across the three downstream tasks. The encoder is evaluated with both frozen and trainable weights during downstream training. For LC, RAMIE, and RARP, procedure-specific pretraining results in segmentation improvements of 13.8\%, 9.5\%, and 36.8\%, respectively, compared to ImageNet initialization with trainable encoder weights. Against ImageNet with frozen weights, improvements are 19.7\%, 25.6\%, and 49.4\%. These findings are in line with the results of Alapatt et al.~\cite{alapatt2023}, who also observe substantial improvements using procedure-specific pretraining datasets. Additionally, our results indicate that training on procedure-specific datasets provides superior performance compared to SurgeNet-Small, despite their comparable sizes.

Moreover, this study indicates that incorporating extra, more heterogeneous data during pretraining further enhances segmentation performance compared to procedure-specific training only. More specifically, training on SurgeNet results in a further improvement of 5.0\%, 5.2\%, and 2.5\% for LC, RAMIE, and RARP, respectively, when encoder weights are trainable, albeit at the expense of a longer training time. Additionally, the results show that SurgeNet pretraining benefits from fine-tuning the encoder in contrast to procedure-specific pretraining, where on both the CholecSeg8k and RARP downstream datasets the highest performance is achieved using a frozen encoder. 

Furthermore, the disparity between ImageNet and SurgeNet initialization is most pronounced on the RARP dataset, which has the smallest amount of labeled training data. This underscores SurgeNet's effectiveness in small-dataset scenarios. However, even on datasets with relatively large labeled data, such as CholecSeg8k, the segmentation model still benefits substantially from SurgeNet pretraining. 
\begin{figure}[h!]
     \centering
     \includegraphics[width=\textwidth]{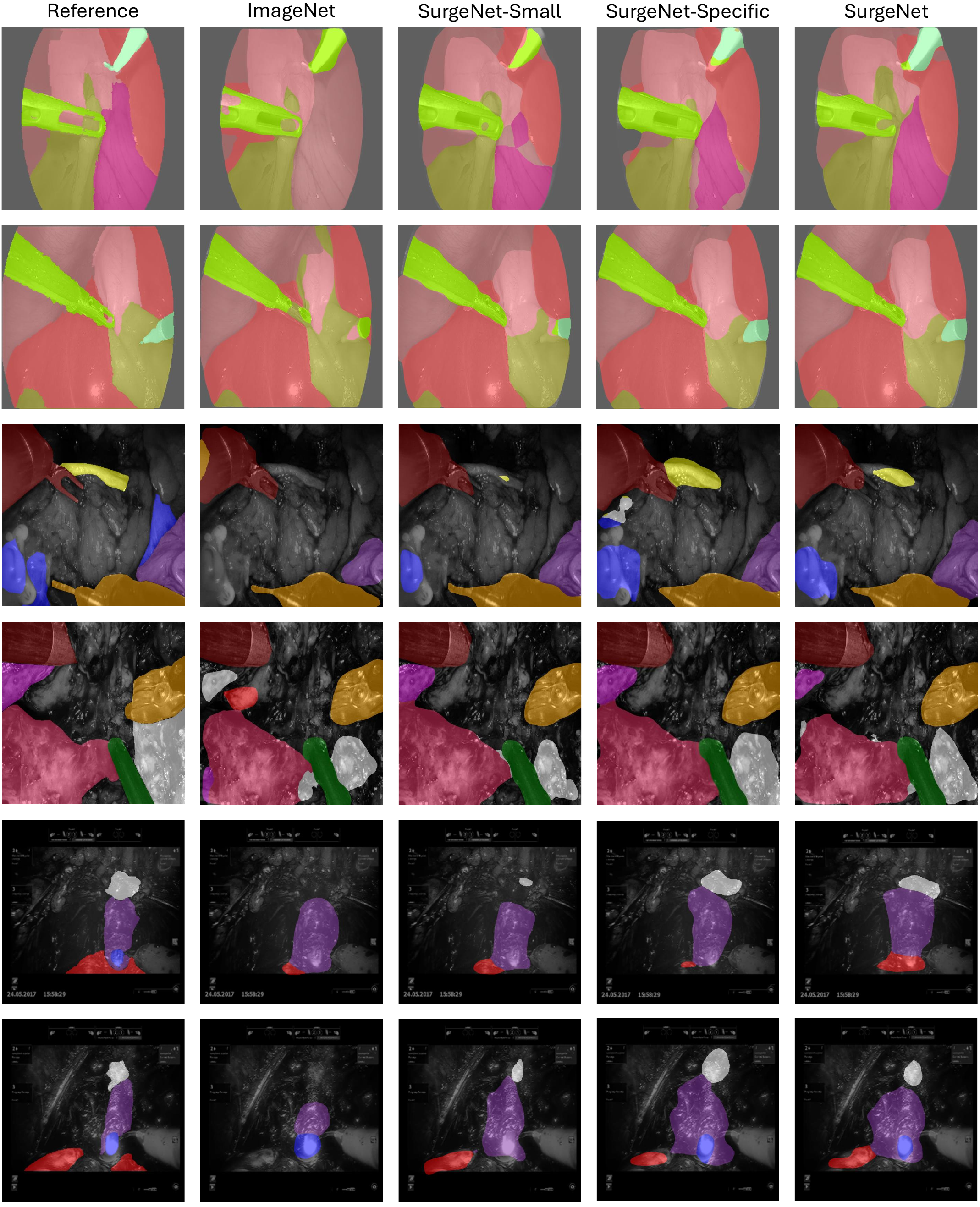}
     \caption{Visual examples from the three downstream test sets. The first two rows display samples from the CholecSeg8k test set, the subsequent two rows show samples from the RAMIE test set, and the final two rows present samples from the RARP test set.}
     \label{fig:visual_results}
     \vspace*{-0.1cm}
\end{figure}
In Figure~\ref{fig:visual_results}, visual results for all three downstream datasets are displayed. The differences between SurgeNet and ImageNet are particularly pronounced in the anatomy classes, while the smallest differences are observed with the segmentation of surgical tools. Small structures e.g. the nerve~(third row, yellow) and the ligated plexus~(last two rows, white) are completely missed by the ImageNet-initialized model. However, both the SurgeNet-specific and SurgeNet-initialized models can accurately detect these important anatomical structures.
\begin{figure}
    \centering
    \includegraphics[width=\textwidth]{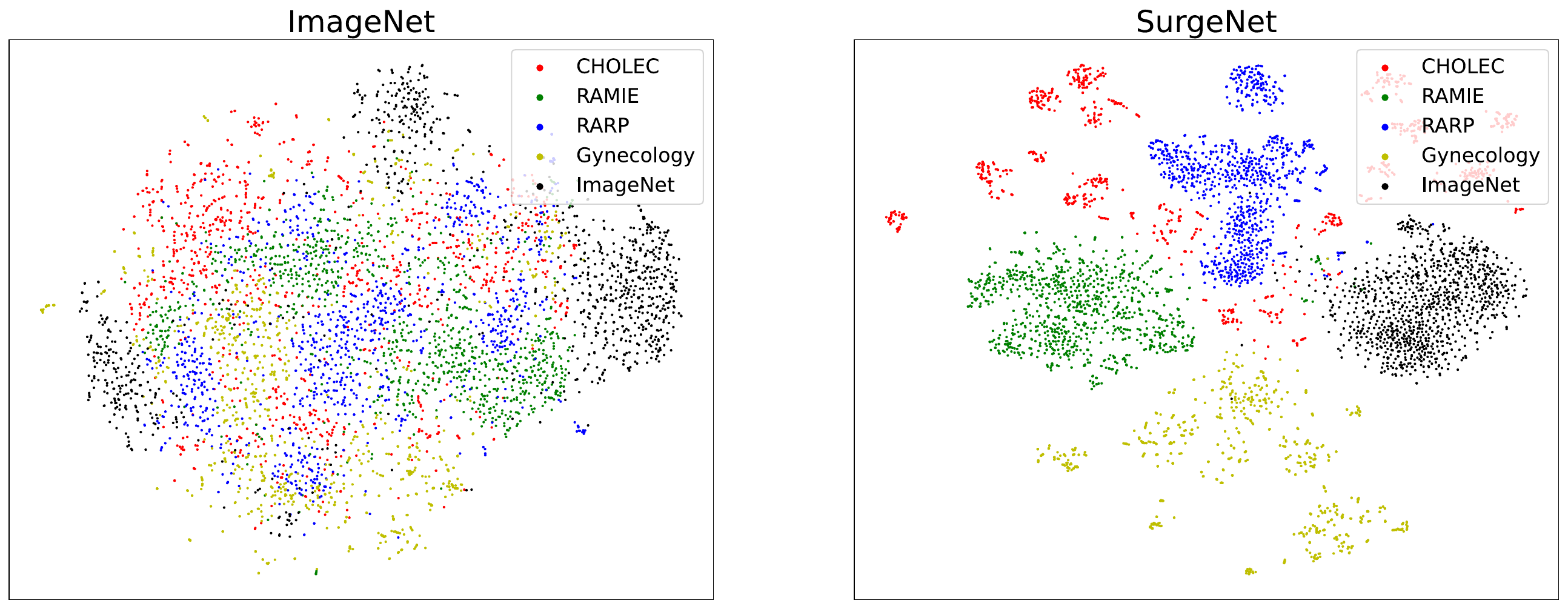}
    \caption{2D-TSNE visualization of learned representations of 1000 randomly sampled images per surgical procedures of SurgeNet and ImageNet.}
    \label{fig:tsne}
    \vspace*{-0.1cm}
\end{figure}

When we consider the learned representations, the 2D-TSNE visualization in Figure~\ref{fig:tsne} indicates that the SurgeNet pretrained CAFormer-S18 can cluster images from the same procedure without any supervision during pretraining. This suggests that the encoder has learned meaningful representations capturing procedure-specific information. Additionally, this capability highlights the potential of the pretrained model to generalize well to unseen data and enhance performance in downstream tasks. These findings underscore the effectiveness of using a heterogeneous and comprehensive dataset like SurgeNet for pretraining, since it enables the model to develop a robust understanding of surgical contexts.
\section{Conclusions and Future work}
This study demonstrates the benefits of increasing data diversity for self-supervised pretraining in surgical computer vision. We construct a large-scale dataset, SurgeNet, containing over 2.6~million surgical video frames from more than seven procedures. When testing on three segmentation downstream datasets, each representing a different surgical procedure, we observe substantial improvements with pretraining on the SurgeNets' procedure-specific datasets, compared to ImageNet by 13.8\%, 9.5\%, and 36.8\%. Incorporating extra data from various procedures by pertaining on the complete SurgeNet, further enhances the results with an additional 5.0\%, 5.2\%, and 2.5\%. The largest performance gains are observed with the smallest downstream datasets, indicating that pretraining on SurgeNet is especially beneficial in small-data scenarios. Additionally, visual examples show the biggest differences on the smallest and often hardest structures to segment. Future work should investigate if pretraining on SurgeNet can advance other applications, like surgical phase or action recognition, where next to temporal feature extraction, spatial feature extraction is also crucial. Although our results show that the model pretrained with SurgeNet can cluster different surgical procedures without supervision, it remains to be seen if pretraining with SurgeNet captures more complex relationships, such as temporal cues between frames. The obtained SurgeNet pretrained weights provide a superior alternative to ImageNet pretrained models for surgical computer vision and are made publicly available at \url{https://github.com/TimJaspers0801/SurgeNet}.

\begin{credits}
    \subsubsection{\ackname} We thank SURF (\url{www.surf.nl}) for the support in using the National Supercomputer Snellius.
\end{credits}
%
%
%
\bibliographystyle{splncs04}
\bibliography{references}

\clearpage

\section{Supplementary Materials}
\begin{figure}
    \centering
    \includegraphics[width=\textwidth]{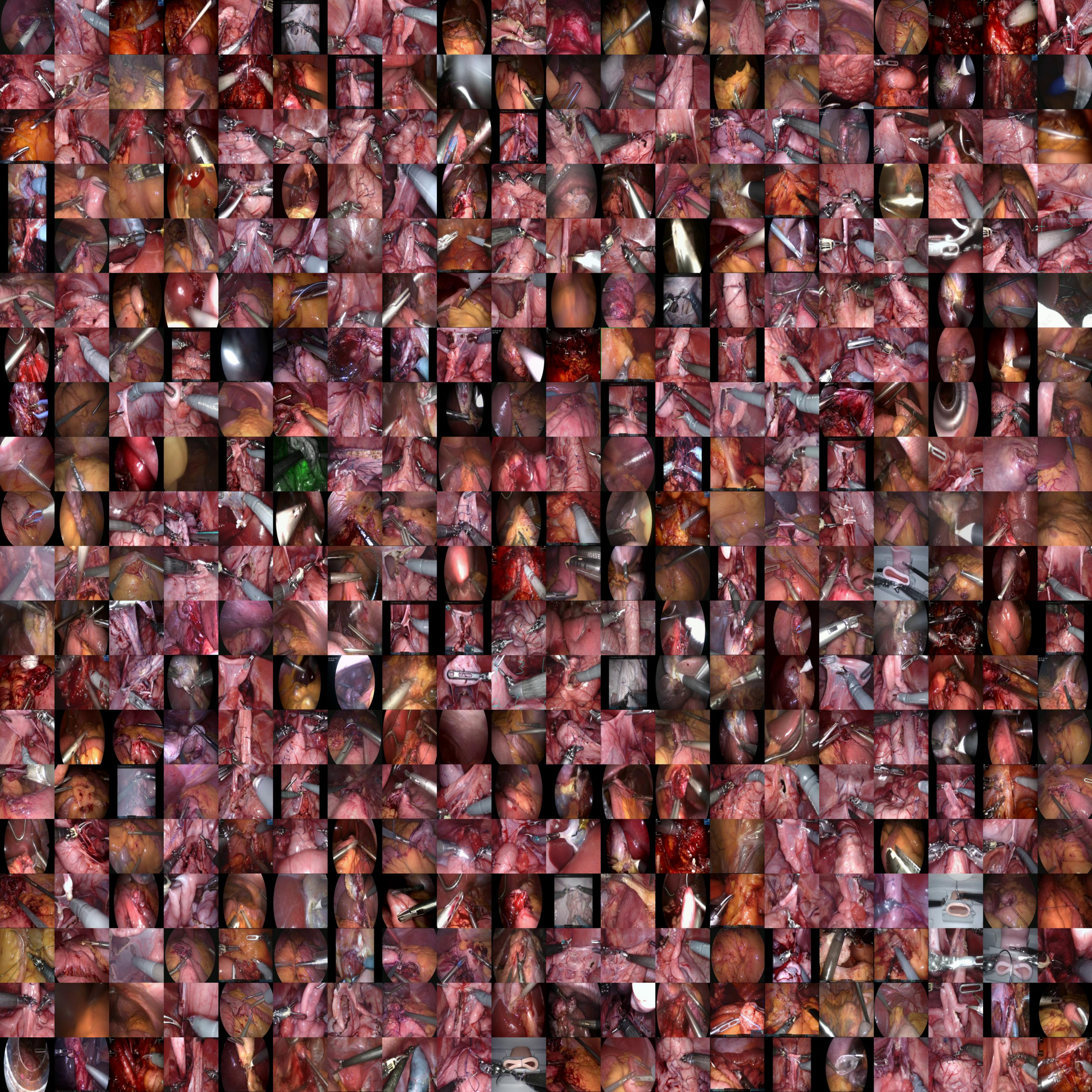}
    \caption{400 randomly sampled images from SurgeNet.}
    \label{fig:surgenet_examples}
    \vspace*{-0.1cm}
\end{figure}

\begin{figure}
    \centering
    \includegraphics[width=\textwidth]{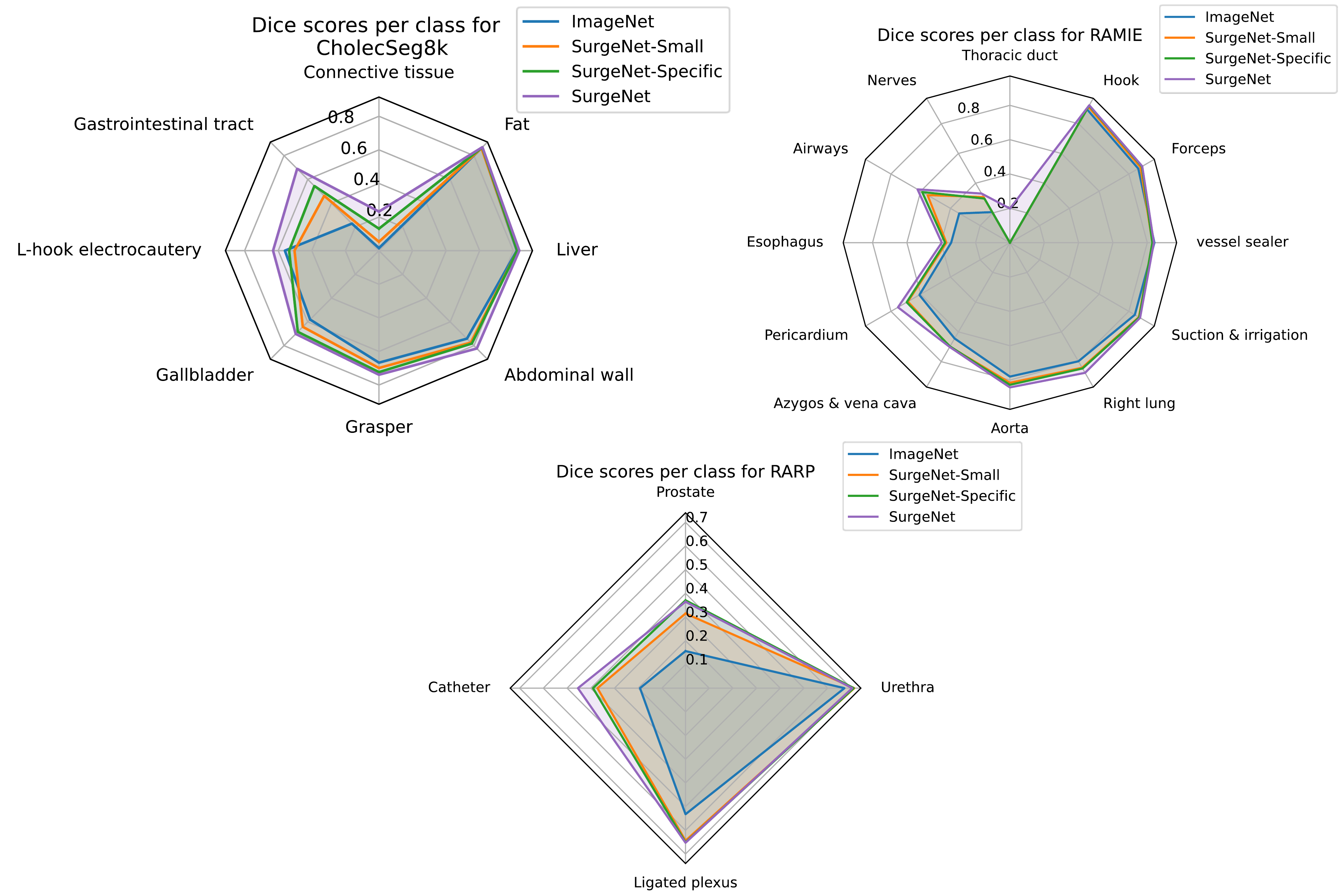}
    \caption{Per-class performance evaluation on three downstream datasets. The top left radar plot displays the results for the CholecSeg8k test set. The top right plot presents the results for the RAMIE test set. The bottom plot illustrates the results for the RARP dataset. The differences are most pronounced for the smaller anatomical structures.}
    \label{fig:surgenet_examples}
    \vspace*{-0.1cm}
\end{figure}

\end{document}